\documentclass[11pt]{article}

\usepackage[utf8]{inputenc}
\usepackage[T1]{fontenc}
\usepackage{amsmath}
\usepackage{amssymb}
\usepackage{authblk}
\usepackage[margin=2.5cm]{geometry}
\usepackage[numbers]{natbib}
\usepackage{hyperref}
\usepackage{siunitx}
\hypersetup{colorlinks=true, linkcolor=blue, citecolor=blue, urlcolor=blue}
\usepackage{graphicx}
\usepackage{subcaption}
\usepackage{tabularx}
\usepackage{booktabs}
\usepackage{xcolor}
\usepackage{array}
\usepackage[table]{xcolor}

\title{\LARGE Vehicle speed dataset for the major European road network derived from Sentinel-2 imagery, 2022–2026}

\author[1,3]{Maciej Adamiak \thanks{Corresponding Author: maciej.adamiak@heigit.org}}
\author[1]{Sascha Fendrich}
\author[1,2]{Julian Psotta}
\author[1,2]{Alexander Zipf}
\author[1]{\authorcr Clemens Langer}
\author[1]{Levi Szamek}
\affil[1]{Heidelberg Institute for Geoinformation Technology (HeiGIT), Germany}
\affil[2]{GIScience Research Group, Institute of Geography, University of Heidelberg, Germany}
\affil[3]{Faculty of Geographical Sciences, University of Lodz, Poland}

\date{}

\begin{document}

\maketitle

\section*{Abstract}

The dataset provides individual vehicle speed observations on European E-roads: motorways, trunk roads, primary and secondary roads, as tagged in OpenStreetMap \texttt{e-road}, for the years 2022-2026. 
Speeds are derived from Copernicus Sentinel-2 Level-2A satellite optical imagery using a processing pipeline that exploits the short, well-characterized acquisition delays between the blue (B02\_10m), green (B03\_10m), and red bands (B04\_10m) of the Sentinel-2 push-broom instrument.
A moving vehicle appears at slightly displaced positions in the three bands, forming a moving echo.
The detected displaced colour-excess peaks are linked into per-vehicle trajectories through a prediction-and-matching procedure.
The resulting displacements are converted into ground speeds using publicly accessible inter-band time delays.
Each record contains the trajectory geometry, per-channel displacements and headings, internal quality indicators, the estimated speed, the acquisition timestamp, and the source Sentinel-2 product identifier.
The dataset is distributed as GeoPackage files, with one record per detected vehicle, and can support studies of traffic patterns, speed behavior, transport modeling, and the calibration of road network attributes at a continental scale.

\section*{Background \& Summary}

Vehicle speed measurements form the basis of any mobility analysis.
Moreover, any mobility analysis that is not linked to real-time or historical traffic data and that estimates realistic travel speeds and the resulting travel times remains a persistent and unresolved challenge.
The majority of topics, such as infrastructure planning, transport modeling, road safety analysis, road traffic management, and humanitarian aid logistics, have historically been limited to heuristically derived maximum permitted speeds for each road type \citep{osm2026}.
Increasingly, vehicle speed measurements also support climate action, for example by modeling emissions from traffic flow; humanitarian aid efforts, such as planning safe evacuation routes or optimizing relief logistics in crisis zones; and tourism, to estimate concentration patterns near reception sites.
Conventional traffic data sources such as inductive loops, roadside radar, and camera installations deliver precise, continuous measurements, but only at fixed, sparsely distributed locations, since each measurement point requires its own installation. 
Floating-car data, speeds reported by navigation devices and connected vehicles, cover road networks more broadly and, because they are collected continuously, support temporal analyses such as recurring congestion patterns and variation by time of day, day of week, or season.
However, their representativeness is bounded by the share of vehicles that actually report. 
Where penetration is low or skewed towards particular driver groups or regions, the resulting sample misrepresents actual traffic conditions, and access to the data is typically restricted by commercial licensing.
Therefore, both sources depend on something external to the road itself — dedicated ground infrastructure in one case, the participation of road users in the other.
Satellite remote sensing removes both dependencies: it provides an independent observation channel with homogeneous spatial coverage, requiring neither ground installations nor data shared by road users.

Optical push-broom satellite instruments like Sentinel-2 don't capture all spectral bands simultaneously. 
Instead, each band is recorded by a separate detector line. 
The satellite's forward motion causes each band to sweep over the same ground location at slightly different times. 
This creates a short interval — a fraction of a second — between when each spectral band images the same point on Earth.
A fast-moving object during that interval is imaged at slightly different positions in each spectral band. 
Therefore, a single scene implicitly contains multiple short, highly localized, multi-frame sequences from which individual velocities can be estimated. 
This effect has been used to measure ocean waves, clouds, aircraft, and road vehicles \citep{kaab2014, heiselberg2019, fisser2022}. 

In previous work, the authors developed a deep learning (DL) method to detect such moving echoes of road vehicles in PlanetScope SuperDove imagery and estimate their velocities. 
The output was validated against drone video footage and GPS trajectories, demonstrating that the method is reliable and can serve multiple purposes in traffic flow analysis and modeling \citep{adamiak2025}. 
However, the study relied solely on a commercial satellite constellation whose imagery is subject to licensing restrictions. 
That constrained both the reproducibility of processing and the redistribution of derived data.
The present dataset transfers the underlying measurement principle to the Copernicus Sentinel-2 mission \citep{drusch2012}, whose imagery is free and openly redistributable.
Moreover, the inter-band time delays of Sentinel-2 have been characterized independently and in detail, including their variation across the field of view and along the orbit \citep{binet2022}. 

The provided dataset was produced with NibbleRGB~\citep{adamiak2026a}, a purpose-built geoprocessing pipeline that combines road geometries from OpenStreetMap (OSM)~\citep{osm2025}, a collaboratively maintained global map database, with Copernicus Sentinel-2 Level-2A surface reflectance imagery~\citep{cdse2026}. 
In OSM nomenclature, a \emph{motorway} is a restricted-access divided highway, a \emph{trunk} road is the most important non-motorway road in a country's network, a \emph{primary} road is the next tier of major road linking large towns, and a \emph{secondary} road is one that is not part of the major routes but forms the national road network~\cite{osm2025}.
The dataset covers these four classes, including their associated link roads such as slip roads and ramps.

Related datasets and tools include truck detections from Sentinel-2~\citep{fisser2022}, generic motion detection from near-simultaneous acquisitions~\citep{kaab2014}, and the authors' PlanetScope speed estimates~\citep{adamiak2025}.
Because the input imagery is freely available from the Copernicus program and the processing pipeline code is open-source, every record in the dataset can be traced back to a specific Sentinel-2 product, regenerated independently, and the whole dataset reproduced.

\section*{Methods}

The dataset was generated with NibbleRGB version 1.3.0, a Python package developed by the authors. 
The pipeline consists of four stages: (i) retrieval and segmentation of road geometries from OSM; (ii) retrieval of Sentinel-2 Level-2A imagery clipped to those road segments; (iii) detection of vehicle echoes and their linkage into per-vehicle trajectories; and (iv) conversion of trajectories into displacement, heading, and speed estimates. 
All processing steps described below correspond directly to the released GitLab source code.

\subsection*{Input data}

\paragraph{OpenStreetMap road network.}

Road geometries (Fig.~1) are retrieved via the Ohsome API V2 at \url{https://api.heigit.org/ohsome-api/v2}, using the \texttt{/extraction/collections} endpoint, which returns the member ways of the requested OSM relations as a GeoParquet collection, clipped to the bounding box of the area of interest.
A unique relation identifier is assigned to each road of interest and passed to the API as an \texttt{id:(relation/<id>, \ldots)} filter.
The code repository includes example configurations for individual roads and regions, as well as the full list of identifiers used for the 2022-2026 production run.
Relations are included in the production list only if their \texttt{route} tag is \texttt{road} and the \texttt{network} is \texttt{e-road}~\cite{eroads2005}.
Although this constraint removed some major roads, mainly in Norway, due to a lack of appropriate tagging, the filter was deliberately retained.
OSM relations are mature and actively maintained, so the selection remains stable and reproducible from a fixed list of relation identifiers.
Additionally, relation member ways are preserved only if they have one of the tags \texttt{motorway}, \texttt{motorway\_link}, \texttt{trunk}, \texttt{trunk\_link}, \texttt{primary}, \texttt{primary\_link}, \texttt{secondary}, or \texttt{secondary\_link}, which is applied server-side through the Ohsome member filter.
Such an approach ensures that the dataset contains only major roads with a total length of \SI{209437}{\kilo\meter}.
The Ohsome API serves the full OSM history, so the retrieved geometries reflect the map state at the timestamp requested in the query, set to the end date of each production run.
The queries for this dataset were executed between 10.07.2026 and 12.07.2026.
An equivalent road network can be retrieved independently by querying the same relation identifiers against the Ohsome API, or by extracting them from a dated OSM archive file covering the requested timestamps.

\begin{figure}[ht]
\includegraphics[width=0.8\textwidth]{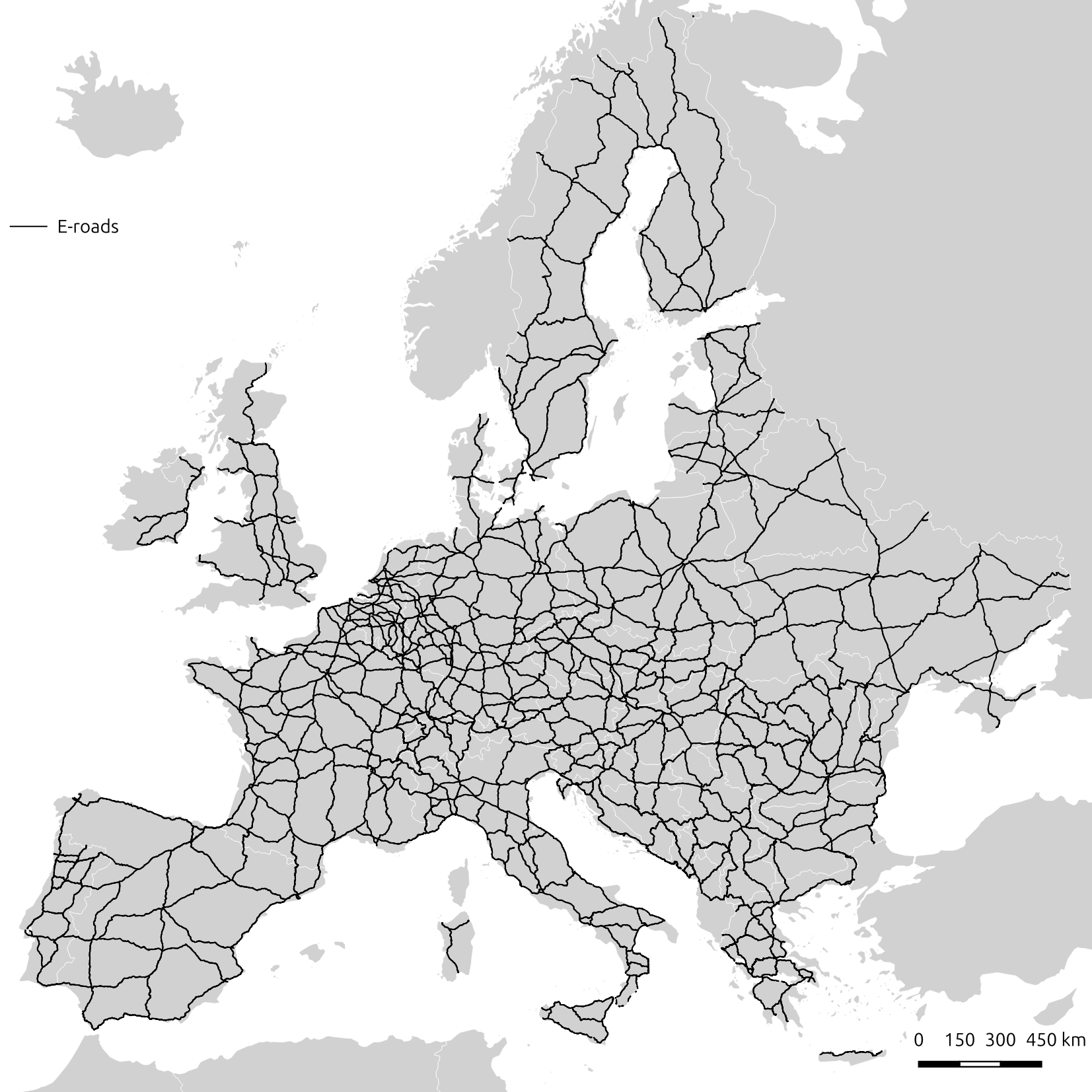}
\centering
\caption{Road geometries (OSM Europe/E-road network)}
\label{fig:e_roads}
\end{figure}

\paragraph{Sentinel-2 imagery.}
Remote sensing imagery is obtained from the Copernicus Data Space Ecosystem through its SpatioTemporal Asset Catalog (\url{https://stac.dataspace.copernicus.eu/v1}) and the Element84 Earth Search (\url{https://earth-search.aws.element84.com/v1}).
The imagery comes from the \texttt{sentinel-2-l2a} collection, which contains Level-2A orthorectified surface reflectance products~\citep{cdse2026}. 
For each road segment, all scenes intersecting its bounding box between 01.06.2022 and 30.06.2026 were queried, and only scenes with reported cloud cover (\texttt{eo:cloud\_cover}) below the configured maximum of 30\% were retained. 
Only the three \SI{10}{\meter} bands of the visible spectrum are fetched from the resource: B02 (blue, $\sim$\SI{490}{\nano\meter}), B03 (green, $\sim$\SI{560}{\nano\meter}) and B04 (red, $\sim$\SI{665}{\nano\meter})~\citep{s2mission2026}.
The pixels are read as unsigned 16-bit digital numbers without rescaling, clipped to the segment polygon, with pixels outside the segment set to a no-data value of zero. 
In addition, the \SI{20}{\meter} scene classification layer (SCL) is fetched and resampled to the \SI{10}{\meter} grid; pixels classified as cloud shadow, medium- or high-probability cloud, thin cirrus, or ice and snow (classes 3, 8, 9, 10 and 11) are masked out and excluded from detection. 
Individual Sentinel-2 products are identified in the dataset by their product identifier.

\paragraph{Inter-band time delays.}

The conversion from pixel peak displacement to speed uses the mean inter-band temporal offsets of the Sentinel-2 multispectral instrument \citep{binet2022}.
The offsets relevant to this dataset are \SI{0.527}{\second} between B02 and B03 and \SI{0.478}{\second} between B03 and B04 (derived as the difference of the published offsets of B03 and B04 with respect to B02), giving a total observation window of \SI{1.005}{\second} per scene. 
Binet~et~al.\ \citep{binet2022} report that these mean values are subject to static and dynamic variations of a few percent across the field of view and along the orbit.
The resulting relative error bound on speed is discussed under Technical Validation.

\subsection*{Road segmentation}

For each OSM road relation, the member way geometries (lines) are clipped to the area of interest (AOI) defined in the configuration files, which is represented as a continent-scale polygon.
Clipped geometries are dissolved and merged into a single connected path.
The path is then projected to the local Universal Transverse Mercator (UTM) zone to enable precise \SI{20}{\meter} buffering on each side.
This step creates a spatial corridor for clipping the remote sensing imagery mask and simultaneously ensures that each road falls correctly within the analyzed region. 
The corridor centerline is derived as its Voronoi-based medial axis (outline densified every \SI{10}{\meter}, result simplified with a \SI{5}{\meter} tolerance).
Buffered corridors with an area exceeding a threshold of \SI{0.5}{\kilo\meter\squared} are split into smaller segments using a recursive algorithm based on buffer centerline delineation.
Segment outlines are simplified with the Douglas--Peucker algorithm \citep{douglas2011} using a \SI{0.5}{m} tolerance.
The centerline part falling within each segment is stored alongside it (layer \texttt{centerlines}) and serves as the heading reference in echo detection.
Each resulting segment is processed independently, which limits memory use and allows scenes to be analyzed in parallel or in a distributed fashion.

\subsection*{Echo detection}

The core of the pipeline is the echo detection algorithm (\href{https://gitlab.heigit.org/traffic-flow-estimator/nibblergb/-/blob/main/nibblergb/analysis/echo.py}{nibblergb/analysis/echo.py}). 
It operates on the three-band image stack from one scene, clipped to a single road segment, with the band acquisition order B02 $\rightarrow$ B03 $\rightarrow$ B04.
The algorithm produces per-vehicle trajectories consisting of one sub-pixel position per band. 
The algorithm consists of two parts: per-band colour-excess extraction and trajectory construction (Fig.~2).

\begin{figure*}[t!]
    \centering
    \begin{subfigure}[t]{0.5\textwidth}
        \centering
        \includegraphics[width=\textwidth]{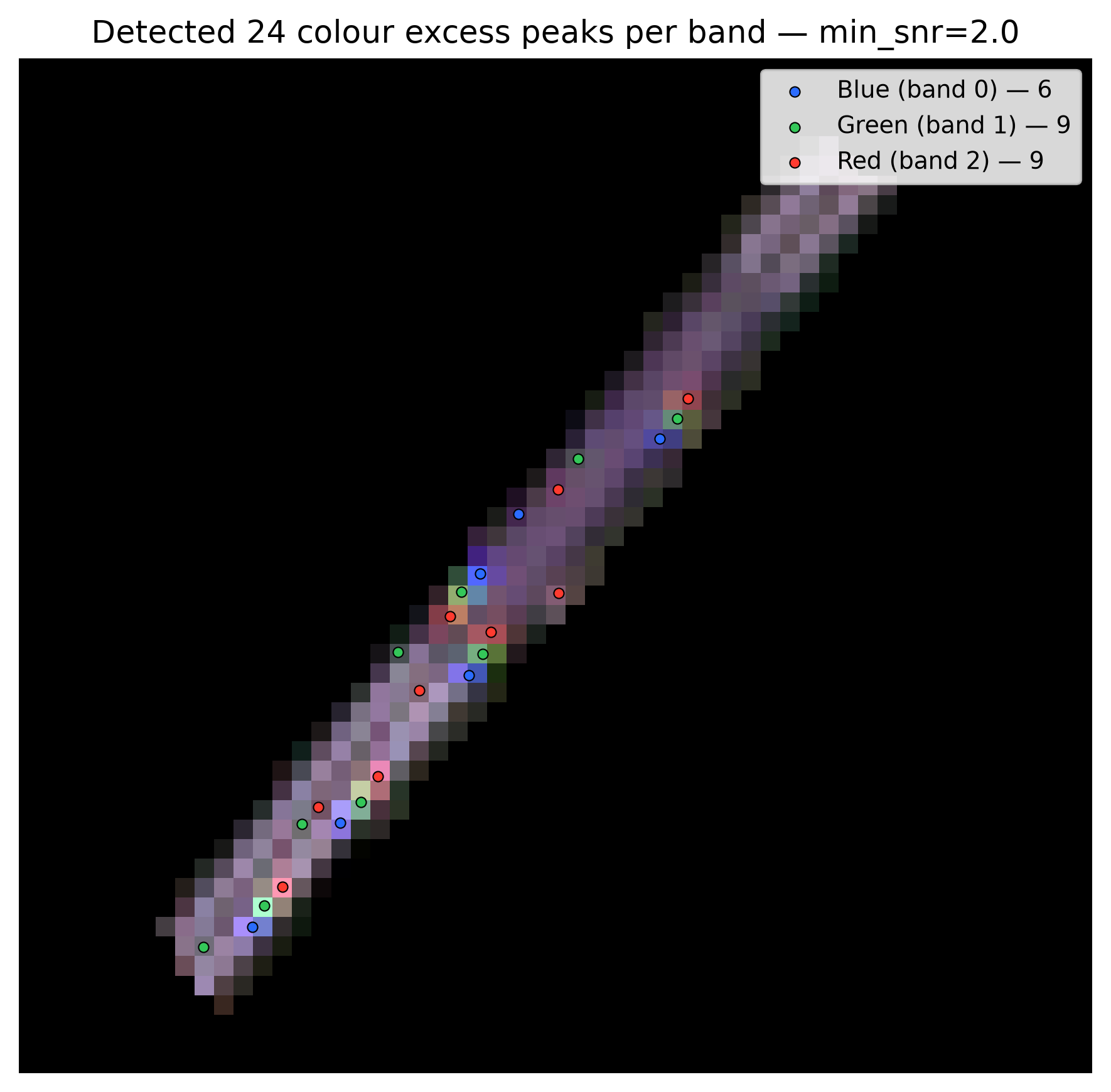}
        \caption{Band peak extraction}
    \end{subfigure}%
    ~ 
    \begin{subfigure}[t]{0.5\textwidth}
        \centering
        \includegraphics[width=\textwidth]{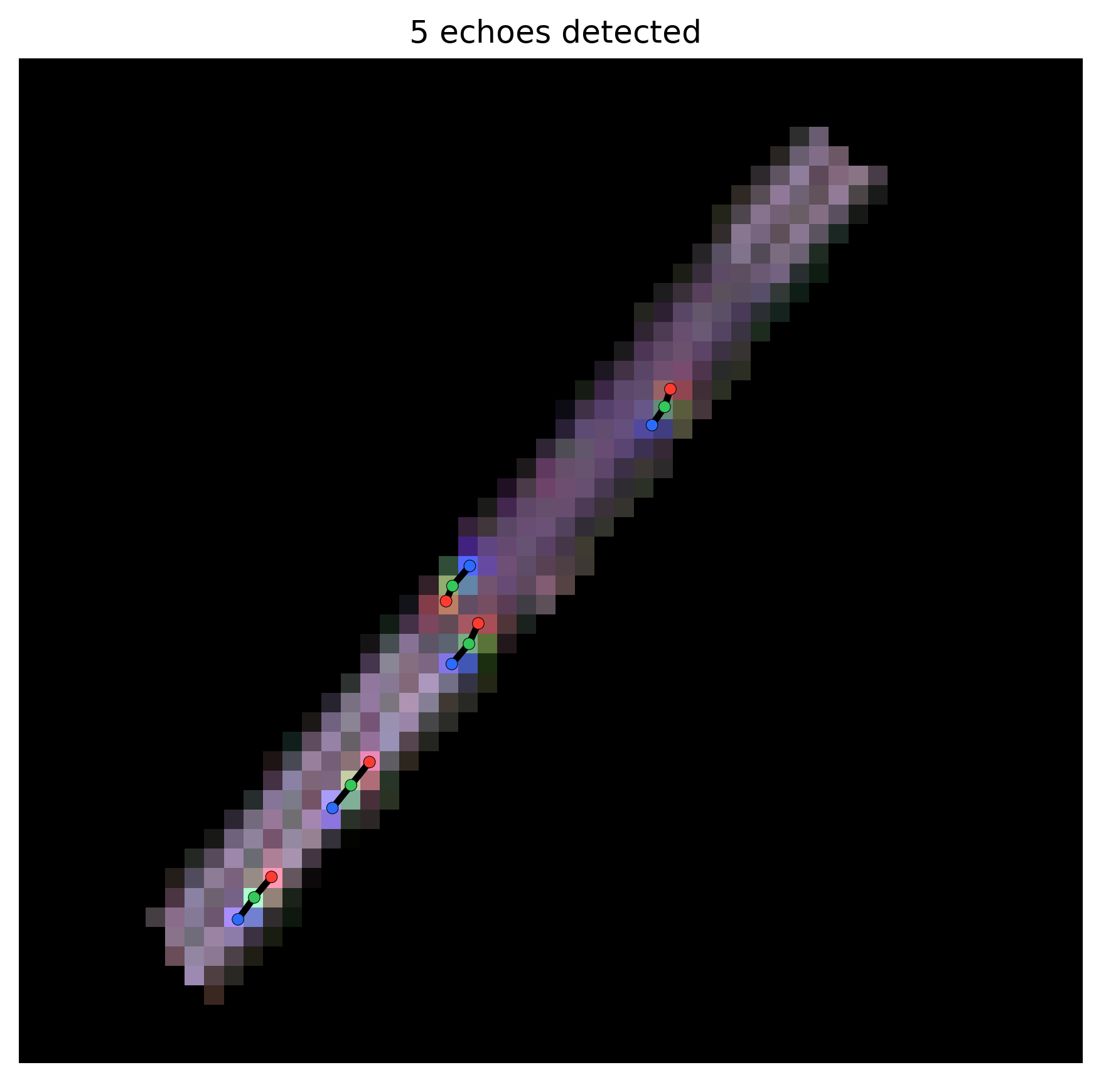}
        \caption{Trajectory extraction}
    \end{subfigure}
    \caption{Moving echo detection on a A1 highway segment in Poland; imagery source: Copernicus Sentinel-2}
    \label{fig:echo_detection}
\end{figure*}

\paragraph{Per-band peak extraction.}
Peaks are extracted separately for each band.
The analysis is restricted to the pixels of the road segment that carry valid (non-zero) data in all three bands, so that peaks are never sought where any band lacks coverage. 
For each band, the steps are as follows: 
(i) \textbf{Compute the colour excess} - the band minus its linear prediction from the mean of the other two bands, with the gain set to the ratio of their standard deviations over the valid pixels; road texture, markings and static objects seen by all bands cancel out, whereas each echo of a moving vehicle remains as an excess of its own band;
(ii) \textbf{Standardize} - the colour excess is expressed as a robust z-score, centred on its median and divided by its median absolute deviation over the valid pixels, so that the thresholds below are noise-relative and stable from tile to tile;
(iii) \textbf{Remove the local background} - a Gaussian-smoothed ($\sigma=2$~pixels) copy of the z-scores, clipped to $\pm2$ so that the echoes do not leak into it, is subtracted and the result is standardized again, which suppresses spatially coherent colour variations of the road surface;
(iv) \textbf{Find the interior} - the valid mask is eroded by a disk of radius $1$~pixel to obtain the interior pixels, away from the edges;
(v) \textbf{Gather peaks} as the interior 4-connected regional maxima that rise at least $1$ above the saddle towards any higher maximum and reach $z>2$ in B02 and B04, and $z>1$ in B03;
(vi) \textbf{Refine to sub-pixel precision} - independently along each image axis, a parabola is fitted through the peak pixel and its two neighbors, and the coordinate is shifted to the parabola vertex, with the shift limited to half a pixel (peaks on the image border keep their integer coordinates).
The output of this stage is one set of sub-pixel peak coordinates per band.

\paragraph{Trajectory construction.}
\begin{figure}[ht]
\includegraphics[width=\textwidth]{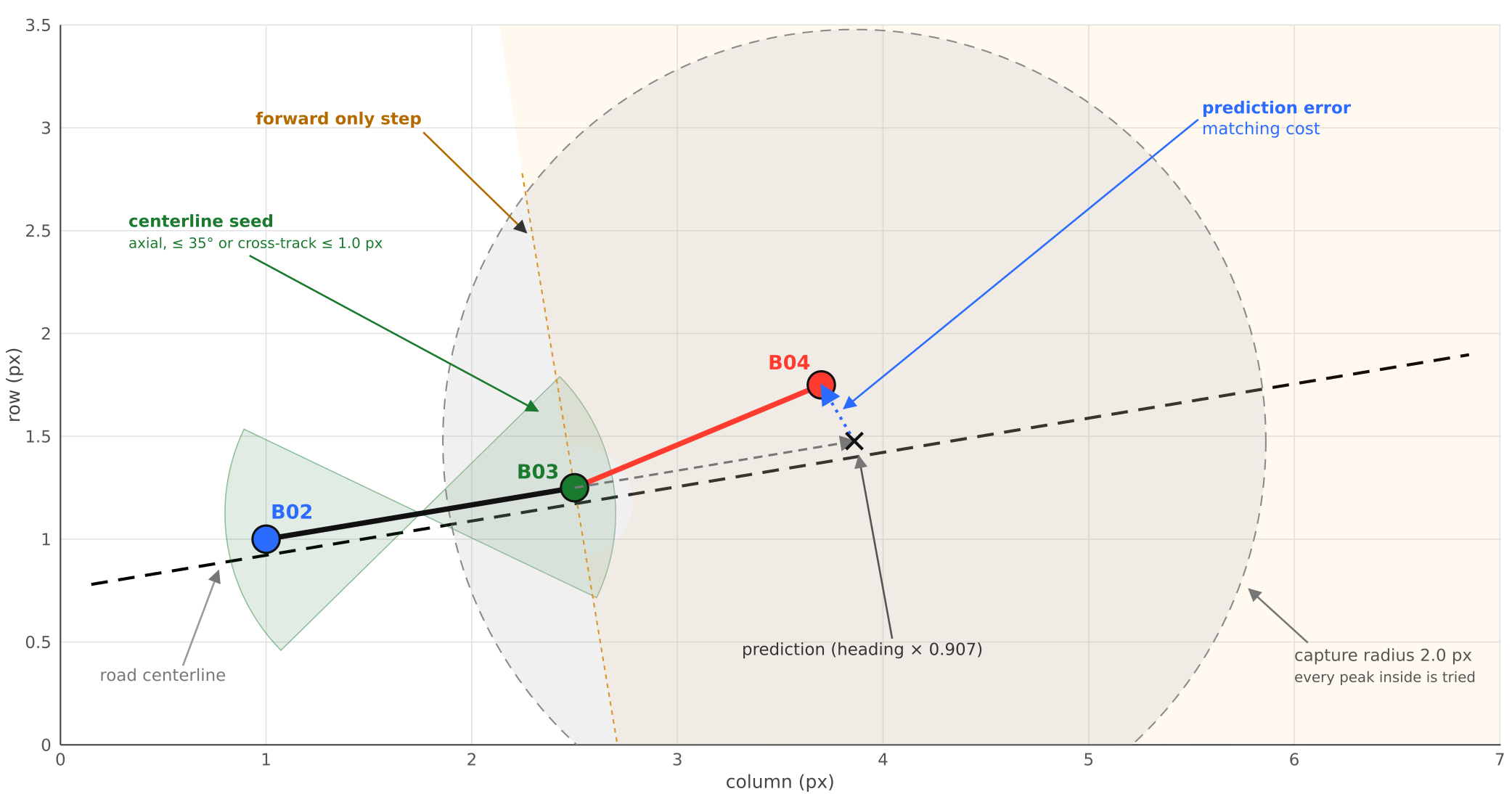}
\centering
\caption{Trajectory construction in image coordinates}
\caption*{\footnotesize Seed displacement between a B02 and a B03 peak (black arrow; admissible length 0.1–3.0px per gap) defines the trajectory heading and, where a road centerline is available, must be near-parallel to it (axial tolerance $35^\circ$, green wedge).
The B04 position is predicted by advancing along the heading scaled by the gap ratio 0.907 (grey cross).
Every B04 peak within the 2.0px capture radius is tried, provided its time-normalized displacement falls within the same bounds, differs from the seed step by at most 1.2px, and points forward.
The prediction error (blue), summed over extension steps, is the matching cost used for conflict resolution.}
\label{fig:trajectory_construction}
\end{figure}

Candidate echoes are formed by systematically linking peaks across the bands in a predefined acquisition order (Fig.~3).
The implementation supports an arbitrary number of bands; for this dataset, three bands are linked.
All displacements are expressed in pixels per band-gap duration.
The procedure for constructing the vehicle trajectories has four stages:

\emph{Trajectory seeding.} Every peak in the B02 band is paired with every peak in the B03 band whose distance from it lies within the admissible per-gap displacement bounds: at least 0.1 and at most 3.0~pixels. 
The vector between the two peaks defines the initial trajectory heading.
Where a road centerline is available, the seed step must additionally be near-parallel to the local centerline direction (evaluated axially, i.e.\ modulo 180\textdegree, over a $\pm1$~pixel window around the point of the centerline nearest to the seed midpoint).
Since a vehicle cannot change heading sharply within the 1.005~s acquisition window, this rejects strongly bent trajectories, while still admitting the heading changes possible on curved link roads.

\emph{Trajectory extension.} For the B04 band, the position is predicted by advancing the current trajectory tip along the heading, scaled by the gap ratio of that step.
The gap ratio is approximately 0.907 (the delay of the second gap, \SI{0.478}{\second}, divided by the delay of the seed gap, \SI{0.527}{\second}), so that a vehicle moving at constant speed is predicted to be at its physically expected position. 
Every detected peak within the capture radius of 2.0 pixels of the prediction is tried as a continuation, so that a seed may branch into several candidate trajectories; a continuation is admitted only if three conditions hold: (i) the time-normalized realized displacement lies within the same per-gap bounds as the seed; (ii) the realized displacement points forward, i.e.\ its angle to the current heading is below 90\textdegree, since a vehicle cannot reverse within the acquisition window; and (iii) the time-normalized realized step length differs from the seed step length by at most 1.2 pixels.
If any condition (i--iii) fails, the candidate trajectory is discarded. 

\emph{Evidence gate.} Each node of a completed candidate contributes the z-score of its own band at the node pixel, reduced by half of the largest positive z-score of the other two bands there and clipped to $[0, 4.5]$; a candidate is retained only if the contributions sum to at least $9$.

\emph{Trajectory acceptance and scoring.} 
Each completed candidate is assigned a matching cost equal to the sum of its prediction errors in pixels; the minimum cost is $0$, so a low cost indicates a smooth, predictable trajectory.
Rather than greedily accepting the cheapest candidates, which lets one candidate consume the peaks of two good ones, every group of mutually conflicting candidates is resolved exactly by a branch-and-bound search for the maximum-weight independent set: the selection with the most trajectories wins, ties going to the lowest total cost.
The search is seeded with the greedy selection and limited to $2\times10^{4}$ nodes per conflict group, beyond which the best selection found so far is kept, so the result is never worse than the greedy one.
The final step involves converting the accepted trajectories from pixel coordinates to the scene's coordinate reference system using the clipped raster's affine geotransform.

\section*{Data Records}
The dataset is hosted at {Zenodo}~\cite{adamiak2026b}, occupies \SI{10.69}{\giga\byte} and is provided as a set of GeoPackage~\citep{gpkg2026} files.
GeoPackage (GPKG) is an open, SQLite-based~\citep{sqlite2026} geospatial data format compatible with common contemporary geographic information system (GIS) software (QGIS~\citep{qgis2026}) and libraries (geopandas~\citep{geopandas2026}, GDAL~\citep{gdal2026}).
The dataset deposit is organized as:
\begin{itemize}
  \item \texttt{eu\_kinematics\_<year>.gpkg} - one file per year, each containing the following layers: 
  \begin{itemize}
  \item \texttt{eu\_kinematics\_<year> -- kinematics} - trajectory geometry and data;
  \item \texttt{eu\_kinematics\_<year> -- centerlines} - estimated road centerlines;
  \item \texttt{eu\_kinematics\_<year> -- segments} - computation work units spatial boundary set;
  \end{itemize}
  \item \texttt{default.yaml} - NibbleRGB production configuration baseline;
  \item \texttt{eu\_<year>.yaml} - NibbleRGB production configuration, OSM relation identifiers per year;
\end{itemize}

The GPKG file structure is as follows (Tab.~1).

\begin{table}[ht!]
\centering
\caption{GeoPackage schema}
\label{tab:gpkg_schema}
\small
\begin{tabularx}{\textwidth}{@{}llcX@{}}
\toprule
\textbf{Field} & \textbf{Type} & \textbf{Unit} & \textbf{Description} \\
\midrule
\rowcolor{gray!15}
\multicolumn{4}{l}{\textbf{Layer \texttt{segments}}} \\
\texttt{fid} & int & --- & Automatically generated feature id \\
\texttt{geometry} & geom & --- & Polygon representing the road segment boundaries \\
\texttt{road\_id} & int & --- & Origin OSM relation id \\
\midrule
\rowcolor{gray!15}
\multicolumn{4}{l}{\textbf{Layer \texttt{centerlines}}} \\
\texttt{fid} & int & --- & Automatically generated feature id \\
\texttt{geometry} & geom & --- & LineString representing the road centerline \\
\texttt{road\_id} & int & --- & Origin OSM relation id \\
\midrule
\rowcolor{gray!15}
\multicolumn{4}{l}{\textbf{Layer \texttt{kinematics}}} \\
\texttt{fid} & int & --- & Automatically generated feature id \\
\texttt{geometry} & geom & --- & LineString with three vertices giving the detected vehicle position in bands B02, B03 and B04, in acquisition order; the line therefore points in the direction of travel \\
\texttt{score} & float & --- & Accumulated matching cost of the trajectory; near 0 represents a more reliable trajectory \\
\texttt{echo\_shift\_0} & float & m & Ground displacement between consecutive band positions (B02$\to$B03), measured in the local UTM projection \\
\texttt{echo\_shift\_1} & float & m & Ground displacement between consecutive band positions (B03$\to$B04), measured in the local UTM projection \\
\texttt{mean\_shift} & float & m & Mean of the two displacement values \\
\texttt{std\_shift} & float & m & Standard deviation of the two displacement values; a large standard deviation indicates an irregular trajectory \\
\texttt{echo\_azimuth\_0} & float & $^\circ$ & Forward azimuth of the B02$\to$B03 gap on the WGS84 ellipsoid, clockwise from north, in the range $[-180, 180]$ \\
\texttt{echo\_azimuth\_1} & float & $^\circ$ & Forward azimuth of the B03$\to$B04 gap on the WGS84 ellipsoid, clockwise from north, in the range $[-180, 180]$ \\
\texttt{mean\_azimuth} & float & $^\circ$ & Circular mean of the two azimuths, in the range $[0, 360)$ \\
\texttt{bearing\_consistency} & float & --- & Resultant length of the two azimuths, between 0 and 1; values near 1 indicate straight-line motion \\
\texttt{speed} & float & m/s & Estimated vehicle speed \\
\texttt{acquisition\_date} & datetime & --- & Sentinel-2 scene acquisition timestamp (Coordinated Universal Time) \\
\texttt{product} & str & --- & Identifier of the source Sentinel-2 Level-2A product, allowing retrieval of the original scene from the Copernicus Data Space Ecosystem \\
\bottomrule
\end{tabularx}
\end{table}

\section*{Technical Validation}

Validation of the dataset rests on four elements: the characterization of the timing model, deterministic tests of the data processing code, internal per-record quality indicators, and plausibility checks of the resulting speed distributions.

\paragraph{Timing model.}
Speed is obtained by dividing the trajectory ground displacement by the fixed mean inter-band offsets (\SI{0.527}{\second} and \SI{0.478}{\second}; total \SI{1.005}{\second}).
Binet~et~al.~\citep{binet2022} characterize deviations of the true delays from these means and identify two static contributors: (i) optical distortion and (ii) satellite altitude.
Together they amount to at most $\pm2\%$ of the tabulated value, with an additional attitude perturbation that contributes a further $0.15\%$ peak-to-peak modulation.
Since speed is inversely proportional to the assumed time base, a relative timing error of $\varepsilon$ produces a relative speed error of the same magnitude; the worst case of $\pm2.1\%$ therefore corresponds to at most $\pm\SI{2.7}{km/h}$ at \SI{130}{km/h} and $\pm\SI{4.3}{km/h}$ at the \SI{205}{km/h} detection ceiling.
This error is systematic for a given detector module and scene position rather than random, so it does not average out across vehicles within a scene.

\paragraph{Unit tests.} The code repository includes a deterministic unit test module covering the components of the data processing pipeline. 
There are 6 test suites available containing 195 tests.
The test coverage is 94\%. 
The following core functionalities are tested: (i) the speed formula - verified against hand-computed values for a range of displacement pairs; (ii) the parabolic sub-pixel refinement, the colour excess and its robust z-score, the prominence-based peak extraction, the evidence gate, and the exact conflict resolution - verified on synthetic arrays with known answers; (iii) end-to-end echo detection - (iiia) verified against a reference Sentinel-2 test image with a known number of echoes, and (iiib) for precision, recall and false positives on synthetic roads covering faint, slow, dense and passing traffic, oblique roads, blurred echoes, unequal band radiometry, coloured road texture, and parked and dark vehicles; (iv) the road segmentation - verified to produce a fixed, expected number of segments for a reference road relation and area of interest.

\paragraph{Internal quality indicators.} Every record carries three indicators that allow users to select conservative subsets without reprocessing the imagery.
\texttt{score} is the accumulated prediction error of the trajectory; a value near 0 indicates motion at nearly constant velocity, whereas large values indicate noisy peak localization.
\texttt{std\_shift} measures the disagreement between the two per-gap displacements, thereby flagging violations of the constant-speed assumption within the \SI{1.005}{\second} window, and \texttt{bearing\_consistency} measures the collinearity of the two gap azimuths.
Since vehicle dynamics over one second are effectively constant, low \texttt{bearing\_consistency} or high \texttt{std\_shift} predominantly indicate localization noise or mis-association rather than true maneuvers, which justifies their use as rejection criteria.

\begin{table}[ht!]
\centering
\caption{Monte-Carlo characterization of the speed and direction estimators ($2\times10^{5}$ simulated trajectories per row; Gaussian per-peak localization noise; admissible displacement bounds of 0.1--3.0~px per gap applied). Speed bias is the mean of the surviving estimates minus the true speed; the rejected share counts trajectories discarded by the displacement bounds.}
\label{tab:monte_carlo}
\small
\begin{tabularx}{\textwidth}{@{}>{\centering\arraybackslash}X>{\centering\arraybackslash}X>{\centering\arraybackslash}X>{\centering\arraybackslash}X>{\centering\arraybackslash}X@{}}
\toprule
\shortstack{\textbf{True speed}\\\textbf{[km/h]}} &
\shortstack{\textbf{Speed error}\\\textbf{[km/h], $1\sigma$}} &
\shortstack{\textbf{Speed bias}\\\textbf{[km/h]}} &
\shortstack{\textbf{Azimuth error}\\\textbf{[$^\circ$], $1\sigma$}} &
\shortstack{\textbf{Rejected}\\\textbf{[\%]}} \\
\midrule
\rowcolor{gray!15}
\multicolumn{5}{l}{\textbf{Localization noise $\sigma_p = 0.25$~px (\SI{2.5}{m} per peak)}} \\
20  & 12.5 & $+17.3$ & 44.7 & 5.2 \\
40  & 12.5 & $+9.3$  & 19.5 & 2.1 \\
70  & 12.8 & $+4.8$  & 10.5 & 0.2 \\
100 & 12.7 & $+3.3$  & 7.3  & 0.0 \\
130 & 12.6 & $+2.4$  & 5.6  & 0.5 \\
160 & 12.1 & $+0.7$  & 4.6  & 9.0 \\
180 & 10.7 & $-3.1$  & 4.1  & 35.1 \\
200 & 8.4  & $-11.6$ & 3.7  & 77.5 \\
\midrule
\rowcolor{gray!15}
\multicolumn{5}{l}{\textbf{Localization noise $\sigma_p = 0.50$~px (\SI{5}{m} per peak)}} \\
20  & 25.5 & $+46.3$ & 72.0 & 1.8 \\
40  & 25.1 & $+33.1$ & 45.0 & 1.6 \\
70  & 24.5 & $+20.1$ & 23.1 & 2.1 \\
100 & 23.9 & $+11.9$ & 15.2 & 6.2 \\
130 & 22.7 & $+4.7$  & 11.6 & 18.8 \\
160 & 20.3 & $-4.6$  & 9.5  & 44.2 \\
180 & 18.1 & $-13.4$ & 8.6  & 65.4 \\
200 & 15.7 & $-24.8$ & 7.9  & 84.1 \\
\bottomrule
\end{tabularx}
\end{table}

\paragraph{Plausibility and coverage checks.} The expected magnitude of the random speed error follows from the sensor geometry: with \SI{10}{m} pixels, the parabolic refinement leaves a per-peak localization uncertainty of approximately 0.25--0.5~px (\SIrange{2.5}{5}{m}), and since the mid-point error largely cancels in the two-gap sum, the per-record speed uncertainty is \SIrange{13}{25}{km/h}.
A Monte Carlo simulation (Tab.~2) confirms this value.
It describes two systematic effects: (i) because localization noise can only lengthen the summed path, low speeds are overestimated (a folded-norm bias of approximately $+9$ to $+46$~km/h at and below \SI{40}{km/h}, which grows with noise); (ii) the admissible displacement range of 0.1--3.0~px per gap restricts detection to approximately \SIrange{7}{205}{km/h}.
Within the displacement range, estimates are effectively unbiased by the filter up to about \SI{130}{km/h} (0.5--19\% of simulated vehicles rejected); above approximately \SI{160}{km/h} the upper bound censors the noise distribution asymmetrically, rejecting 9--84\% of simulated vehicles between \SI{160}{} and \SI{200}{km/h} and biasing the surviving estimates low by up to \SI{25}{km/h}, consistent with the underestimation of the highest speeds reported for band-displacement methods~\citep{adamiak2025}.
Direction estimates show the complementary behavior: the azimuth error is approximately \SIrange{3.7}{15.2}{\degree} ($1\sigma$) at and above \SI{100}{km/h} but grows to \SIrange{19.5}{72.0}{\degree} at and below \SI{40}{km/h}, where the displacement approaches the localization noise.

\paragraph{Heading agreement with the road centerline.} The road centerline provides an independent geometric reference for the direction of travel: a detected vehicle must move along the road it occupies, so \texttt{mean\_azimuth} should agree with the local centerline bearing modulo $180^\circ$, the two admissible travel directions. 
Crucially, this agreement is never enforced on the accepted trajectories.
The centerline is used in the pipeline only as an optional reference in the axial alignment test of the seed step during echo detection, so the heading residual remains a genuinely out-of-sample check.

Across 1345000 sampled records, the median absolute heading residual is $6.91^\circ$, with a $1\sigma$ dispersion of $10.24^\circ$.
The dispersion decreases from $16.49^\circ$ below \SI{40}{km/h} to $8.18^\circ$ above \SI{100}{km/h}, reproducing the trend predicted in Table~2.
Matching the observed dispersion to the simulated curves yields an effective localization noise of $\sigma_p \approx 0.24$px.
Since the azimuth and speed errors are driven by the same noise term, this measurement constrains $\sigma_p$ directly from production imagery and thereby corroborates the per-record speed uncertainty reported above.

\paragraph{Dataset overview.} 
\begin{figure}[ht]
    \centering
    \includegraphics[width=\textwidth]{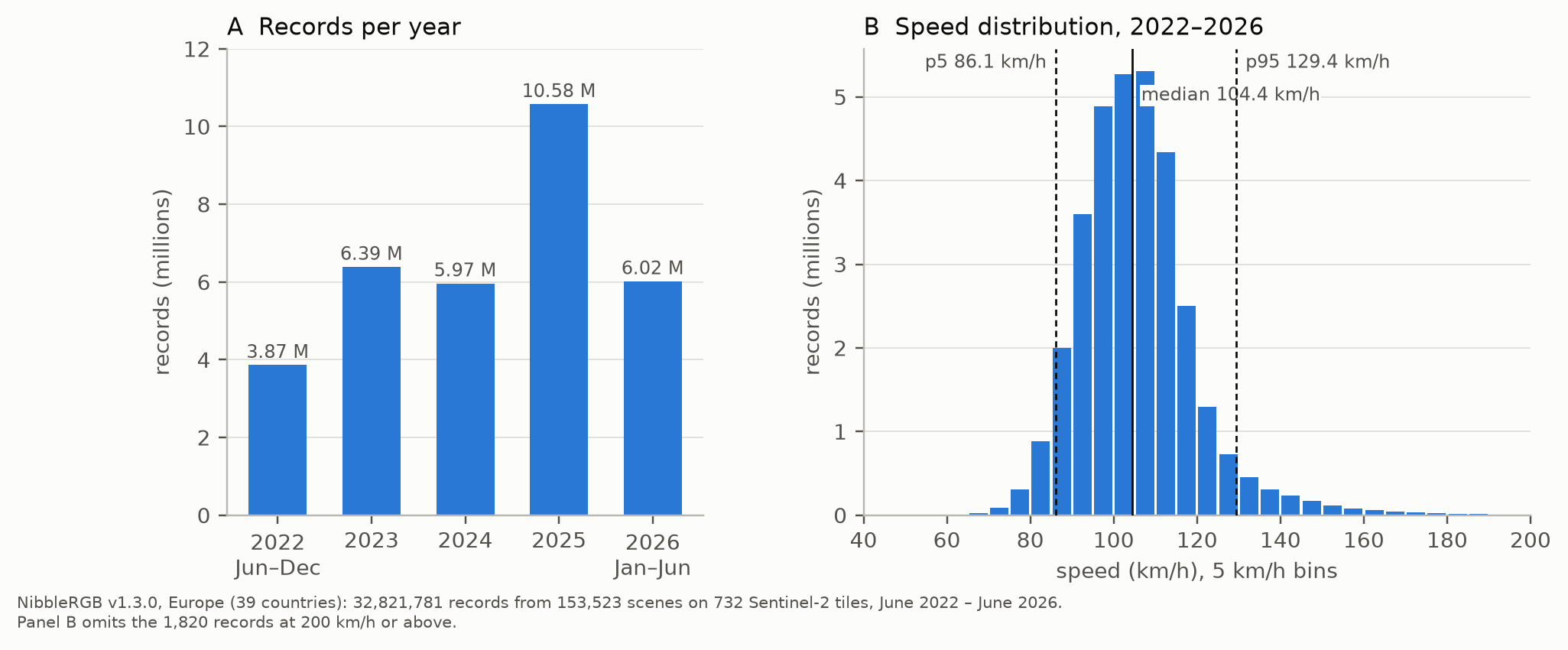}
    \caption{Records per year (A) and their speed distribution (B) for Europe from June 2022 to June 2026 generated with NibbleRGB v1.3.0.}
    \label{fig:v130-data-contents}
\end{figure}
The dataset generated with NibbleRGB contains 32.82 million speed records, detected across 153,523 Sentinel-2 scenes on 732 tiles in 39 European countries (Fig.~4A).
The median speed is 104.4 km/h and 90\% of the records fall between 86.1 km/h and 129.4 km/h (Fig.~4B).

\section*{Usage Notes}

Several properties of the records matter for their interpretation.
Each record is an independent, single-scene observation of a single vehicle.
The same physical vehicle is not tracked across scenes or imagery acquisition days, although the same vehicle can occur in different scenes due to Sentinel-2 tile grid overlap.
In the case of approach traffic volume estimation, aggregating results for overlapping trajectories over time could be considered a feasible approach. 
Because Sentinel-2 operates in a sun-synchronous orbit with a mean local solar time of 10:30 at the descending node \citep{binet2022}, the dataset captures only late-morning traffic and does not capture peak-hour conditions. 
Users should also expect the known limitations of band-displacement speed estimation: reduced sensitivity to vehicles with low contrast against the road surface, no detection of vehicles darker than it, a lower detection bound imposed by the minimum displacement threshold, degraded completeness under haze or partial cloud, and a tendency to underestimate the highest speeds \citep{adamiak2025}.
Further, because the method relies on inter-band displacement, stationary and slow-moving vehicles are not recorded, so the dataset cannot be used to identify congestion or queuing.
For applications that require conservative subsets, filtering on \texttt{score}, \texttt{std\_shift}, and \texttt{bearing\_consistency} is recommended.

\section*{Data Availability}

The dataset is hosted at Zenodo~\cite{adamiak2026b} and conforms to the following licensing rules:
(i) The kinematics GPKG layer is available under the CC BY 4.0 license as derived from Copernicus imagery;
(ii) The GPKG layers created from road geometries (segments and centerlines) are available from OSM under the Open Database License (ODbL);
(iii) The referenced Sentinel-2 Level-2A products are freely available from the Copernicus Data Space Ecosystem. 

\section*{Code Availability}

The NibbleRGB source code (version 1.3.0) used to produce this dataset is available at \url{https://gitlab.heigit.org/traffic-flow-estimator/nibblergb} under the AGPL-3.0 license, together with the production configuration files, the test suite, and the reference test data. 
The code can be run on any machine with internet access, without additional setup. 
However, the preferred approach is to deploy it on a machine located in the same region as the Sentinel-2 data catalog: \texttt{us-west-2} (Oregon) on AWS for  Element84, or \texttt{eu-central-1} (Frankfurt) for Copernicus.
To take full advantage of the pipeline's heavy parallelization, more than one CPU must be used (consider a \texttt{c7gn.xlarge} instance to reduce the cost and optimize computation time).

\section*{Funding}
This work was supported by core funding from the Klaus Tschira Stiftung (KTS, Germany). 
No specific grant number applies to this funding.

\end{document}